\documentclass[letterpaper]{article}
\usepackage[dblblindworkshop, final]{neurips_2025}

\usepackage[utf8]{inputenc}
\usepackage[T1]{fontenc}
\usepackage{times}
\usepackage{helvet}
\usepackage{courier}
\usepackage[hyphens]{url}
\usepackage{graphicx}
\usepackage{hyperref}  
\usepackage{natbib}
\usepackage{caption}
\usepackage{microtype}
\usepackage{amsmath}
\usepackage{amsfonts}
\usepackage{booktabs}
\usepackage{float}
\usepackage{xcolor}
\usepackage{enumitem}
\setlist[itemize]{topsep=0pt, partopsep=0pt, parsep=0pt, itemsep=4pt}

\makeatletter
\newcommand{\samethanks}[1][\value{footnote}]{\footnotemark[#1]}
\makeatother

\usepackage{inconsolata}
\usepackage[most]{tcolorbox}
\tcbuselibrary{listingsutf8}
\newtcolorbox{promptbox}[2][]{
  colback=blue!5!white, 
  colframe=blue!40!white,
  fonttitle=\bfseries, 
  title=#2, 
  #1
}

\title{Reasoning Relay: Evaluating Stability and Interchangeability of Large Language Models in Mathematical Reasoning}

\workshoptitle{Socially Responsible and Trustworthy Foundation Models (ResponsibleFM)}

\author{
  \normalfont
  \hspace{-0.25cm}
  \begin{tabular}{ccc}
    \textbf{Leo Lu}$^{1}$\thanks{Equal Contribution} & \textbf{Jonathan Zhang}$^{2}$\samethanks & \textbf{Sean Chua}$^{3}$\samethanks \\
    Pennsylvania State University & Binghamton University & University of Toronto\\
    \texttt{lbl5561@psu.edu} & \texttt{jzhang78@binghamton.edu} & \texttt{seaneugene.chua@mail.utoronto.ca} \\[1em]
    \textbf{Spencer Kim}$^{4}$ & \textbf{Kevin Zhu}$^{5}$\thanks{Corresponding Author} \thanks{Senior Author} & \textbf{Sean O'Brien}$^{6}$\samethanks \\
    UC Berkeley & Algoverse & Algoverse \\
    \texttt{spencer\_kim@berkeley.edu} & \texttt{kevin@algoverse.us} & \texttt{2000.seano@gmail.com} \\[1em]
    & \textbf{Vasu Sharma}$^{7}$\samethanks & \\
    & Algoverse & \\
    & \texttt{sharma.vasu55@gmail.com} & \\
  \end{tabular}
}

\begin{document}
\maketitle
\begin{abstract}
Chain-of-Thought (CoT) prompting has significantly advanced the reasoning capabilities of large language models (LLMs). While prior work focuses on improving model performance through internal reasoning strategies, little is known about the interchangeability of reasoning across different models. In this work, we explore whether a partially completed reasoning chain from one model can be reliably continued by another model, either within the same model family or across families. We achieve this by assessing the sufficiency of intermediate reasoning traces as transferable scaffolds for logical coherence and final answer accuracy. We interpret this interchangeability as a means of examining inference-time trustworthiness, probing whether reasoning remains both coherent and reliable under model substitution. Using token-level log-probability thresholds to truncate reasoning at early, mid, and late stages from our baseline models, Gemma-3-4B-IT and LLaMA-3.1-70B-Instruct, we conduct continuation experiments with Gemma-3-1B-IT and LLaMA-3.1-8B-Instruct to test intra-family and cross-family behaviors. Our evaluation pipeline leverages truncation thresholds with a Process Reward Model (PRM), providing a reproducible framework for assessing reasoning stability via model interchange. Evaluations with a PRM reveal that hybrid reasoning chains often preserve, and in some cases even improve, final accuracy and logical structure. Our findings point towards interchangeability as an emerging behavioral property of reasoning models, offering insights into new paradigms for reliable modular reasoning in collaborative AI systems.
\end{abstract}

\section{Introduction}

Chain of Thought (CoT) prompting emerged as powerful mechanism to improve the reasoning capabilities of large language models (LLMs) by encouraging intermediate structured reasoning steps before arriving at a final answer \citep{wei2023chainofthoughtpromptingelicitsreasoning}. Previous work has explored how CoTs improve individual model performance even in zero-shot settings \citep{kojima2023largelanguagemodelszeroshot, zhang2022automaticchainthoughtprompting, jin2024zeroshotchainofthoughtreasoningguided}. More recently, \citet{hebenstreit2024comparison} examined the transferability of entire CoT sequences by evaluating whether rationale prompts discovered on one model could generalize reasoning strategies across a range of models and tasks. However, it remains unclear to what extent reasoning trajectories are interchangeable when only partially reused. In light of this, our aim is to answer the central research question: \textit{To what extent can the modular decomposition of complex mathematical reasoning tasks enhance the zero-shot performance and interpretability of Large Language Models, when utilizing a collaborative framework that includes both intra-family and cross-family LLMs?}

    In this work, we investigate the process-level interchangeability in language model reasoning by evaluating how well different models can continue the CoT of another's midstream. We begin with full CoT traces generated by a strong base model (e.g., Gemma-3-4B-IT and LLaMA-3.1-70B-Instruct), recording token-level log-probabilities to guide strategic truncation at $25\%$, $50\%$, and $75\%$ of the cumulative log-probability, capturing early, mid, and late stages of reasoning based on informativeness. From these truncated points, alternative models (including those from different families or architectures) are tasked with continuing the reasoning process using only truncated intermediate steps as input We then assess not only accuracy, but also the coherence, semantic alignment, and logical consistency of the full reasoning chain, by using a Process Reward Model (PRM) trained to evaluate multi-step mathematical reasoning performance. Ultimately, our aim is to characterize how steady transferability depends on truncation point, model pairing, and reasoning domain, yielding clearer interpretations into the dynamics of CoT continuation success that goes beyond final answer accuracy.
    
    Whereas prior work has explored how CoT prompting improves reasoning within individual models \citep{wei2023chainofthoughtpromptingelicitsreasoning}, whether reasoning can be interchanged across models mid-process remains largely unexamined. 

    We provide compelling early evidence that such a handoff is often successful within the same model family. We show that a partially completed CoT from a strong model, such as Gemma-3-4B-IT, can often be continued by another model of similar or lesser capacity within the same family. By leveraging log-probability-based truncation and PRM-based scoring, we found that these hybrid trajectories maintain high coherence and correctness with minimal loss in reasoning quality.
    
    We found that this practice may not be suitable for all cross-family continuation pairings, as some unreliably preserve quality and coherence of the reasoning chain in our experimentation. Our findings expose distinctions across different model architectures and introduce a promising new paradigm for collaborative reasoning, where high-capacity models can be reserved for the most uncertain portions of a problem, allowing lighter models to reliably finish the remainder of the task.

\section{Related Works}

LLMs generate responses by autoregressively predicting outputs based on the preceding context, which is learned during pre-training \citep{openai2024gpt4technicalreport}. As a result, their output can fluctuate even when prompted with identical inputs, introducing variability in reasoning trajectories \citep{amatriain2024promptdesignengineeringintroduction}. This, coupled with the absence of structured reasoning mechanisms, often leads to inconsistencies in multistep logical inference. Consequently, assessing the reliability and soundness of their reasoning becomes increasingly complex and therefore requires a more thorough examination \citep{Wang2024Qstar}.

To address these limitations, the concept of CoT prompting was introduced in \citet{wei2023chainofthoughtpromptingelicitsreasoning}, demonstrating that instructing LLMs to reason step-by-step significantly improves performance on complex tasks. In this approach, LLMs are prompted to generate a series of short statements that mimic the logical process a person might use to solve a problem. Experiments revealed that CoT prompting enables models to achieve strong results in tasks of arithmetic, commonsense, and symbolic reasoning \citep{wei2023chainofthoughtpromptingelicitsreasoning}.

In an effort to enhance LLM reasoning abilities with CoT prompting, \citet{wang2023selfconsistencyimproveschainthought} introduces self-consistency to replace the single greedy decoding path in traditional CoT prompting \citep{wei2023chainofthoughtpromptingelicitsreasoning}. Their method samples a variety of reasoning paths and identifies the most consistent answer by marginalizing across all possibilities \citep{wang2023selfconsistencyimproveschainthought}. Beyond improving accuracy, this approach highlights the inherent diversity of reasoning paths within a single model, suggesting that multiple valid chains of reasoning can coexist.

Initiatives have also been put forward to extend and refine CoT prompting. Unlike traditional CoT where each step is independent, Least to Most prompting breaks difficult problems into sequential sub-problems where the outputs of previous steps are the inputs for the next \citep{zhou2023leasttomostpromptingenablescomplex}. Moreover, recent efforts have examined the effects of partial or truncated CoT on model outputs. \citet{lanham2023measuringfaithfulnesschainofthoughtreasoning} measure faithfulness by truncating generated CoT at various points and re-prompting the model with only the partial reasoning.

Past research has indirectly measured the reasoning ability of LLMs by evaluating them on downstream reasoning tasks such as question answering or multi-hop inference \citep{huang2023reasoninglargelanguagemodels}. Though, relying on the accuracy of the end task or the success rates is not indicative of step-by-step reasoning capability. \citet{huang2023reasoninglargelanguagemodels} also explain that current performance measures mix reasoning ability with task knowledge, resulting in reasoning that cannot be evaluated in isolation. To resolve this, subsequent work \citep{nguyen2024directevaluationchainofthoughtmultihop} aims at reasoning process analysis directly, testing for logical coherence of individual steps, which provides more straightforward methods of reasoning quality evaluation.

LLMs frequently make errors when solving mathematical problems step-by-step, making it essential to identify where the errors occurred during the reasoning process \citep{zheng2025processbenchidentifyingprocesserrors}. As a result, PRMs have been developed as a direct solution to the shortcomings of traditional indirect evaluation methods, which only assess final answers. PRMs are specifically designed to evaluate the correctness of each individual reasoning step, providing feedback that helps guide policy models toward more accurate and reliable mathematical reasoning \citep{zheng2025processbenchidentifyingprocesserrors,zhang2025lessonsdevelopingprocessreward}. PRMs output a score or probability that represents the model’s confidence that the reasoning step is logically sound and contributes productively to problem resolution.

\section{Methodology}
\label{sec:methodology}
\label{headings}
We introduce a novel chain-splitting approach grounded in cumulative log-probability, whereby complete solutions are truncated at points of varying model confidence from an initial baseline model and then continued by a second continuation model. The methodology proceeds in three components: (\hyperref[sec:reasoning-chain]{Section~\ref*{sec:reasoning-chain}}) reasoning chain generation, (\hyperref[sec:chain-truncation]{Section~\ref*{sec:chain-truncation}}) chain truncation via cumulative log-probability, and (\hyperref[sec:model-interchange]{Section~\ref*{sec:model-interchange}}) model interchange protocols.

\subsection{Reasoning Chain Generation}
\label{sec:reasoning-chain}
We use an initial model to generate complete reasoning chains for each problem in the test set. Each generation is performed with temperature set at $0.7$, allowing a moderate degree of stochasticity in token sampling while still favoring high-probability continuations. Let the complete output chain be a sequence of tokens 
\( r = \{t_1, t_2, \ldots, t_n\} \), 
with corresponding log-probabilities 
\( \{\ell_1, \ell_2, \ldots, \ell_n\} \). 
We compute the cumulative log-probability up to position \( i \) as 
\( L_i = \sum_{j=1}^{i} \ell_j \).

This sequence 
\( \{L_1, L_2, \ldots, L_n\} \) 
defines the internal flow of confidence of the model throughout the reasoning process.

\subsection{Chain Truncation via Log-Probability Thresholding}
\label{sec:chain-truncation}
To identify semantically meaningful split points in the chain, we define three thresholds based on the total log-probability \( L_n \):

\begin{itemize}
    \item \textbf{25\% truncation}: first index \( i \) such that \( L_i \geq 0.25 L_n \)
    \item \textbf{50\% truncation}: first index \( i \) such that \( L_i \geq 0.50 L_n \)
    \item \textbf{75\% truncation}: first index \( i \) such that \( L_i \geq 0.75 L_n \)
\end{itemize}

For each threshold \( \alpha \in \{0.25, 0.50, 0.75\} \), we extract the prefix \( r_{1:k} \), where
\[
k = \min\{i : L_i \geq \alpha L_n\}.
\]
This results in three partially completed reasoning traces per problem, each grounded in the model’s own internal confidence progression.

\subsection{Model Interchange Protocol}
\label{sec:model-interchange}
Each truncated prefix is combined with a consistent CoT template meant for interchange that includes the original question, and the resulting prompt is provided to a secondary continuation model (further details in \hyperref[app:prompting]{Section~\ref*{app:prompting}}). We consider both intra-family and cross-family model pairings more precisely defined in \hyperref[sec:model-selection]{Section~\ref*{sec:model-selection}}. Each continuation model generates a single completion for each prefix using a temperature of $0.7$, introducing controlled randomness to reflect typical sampling conditions while preserving coherence. These continuations are concatenated with the original prefix to form hybrid reasoning chains, which are then run through post-processing to extract the final answer using simple rule-based extraction.

All in all, for each problem instance, we obtain: one complete chain from the baseline generator, and multiple hybrid chains resulting from different continuation models and truncation depths (details in \hyperref[sec:model-selection]{Section~\ref*{sec:model-selection}}).

\section{Experimental Setup}
\label{sec:experimental-setup}

We now outline the experimental conditions under which our chain-splitting framework was evaluated. This includes: (\hyperref[sec:dataset-selection]{Section~\ref*{sec:dataset-selection}}) the dataset selected to benchmark reasoning difficulty and domain coverage, (\hyperref[sec:model-selection]{Section~\ref*{sec:model-selection}}) the models used for initial generation and continuation, and (\hyperref[sec:eval-metrics]{Section~\ref*{sec:eval-metrics}}) the metrics employed to quantify the quality of reasoning, compatibility, and the impact of performance on model interchanges.

\subsection{Dataset Selection}
\label{sec:dataset-selection}
An extensive dataset was carefully selected to capture a range of reasoning complexities and domain-specific scenarios.
\begin{itemize}
\item MATH \citep{hendrycks2021measuringmathematicalproblemsolving}: consists of 12,500 high-school and college-level mathematical problems that span diverse topics and demanding multi-step solutions, providing rigorous testing to evaluate advanced mathematical reasoning and generalization. 
\end{itemize}

For our experiments, we evaluated models exclusively on the test splits of the MATH dataset, consisting of $5,000$ questions.

\subsection{Model Selection and Configuration}
\label{sec:model-selection}
We adopt Qwen2.5-PRM \citep{zheng2025processbenchidentifyingprocesserrors} as our primary Process Reward Model, due to its fine-tuning on structured multi-step mathematical datasets such as PRM800K \citep{prm800k} and Math-Shepherd \citep{mathshepherd}. Qwen2.5-PRM is an instruction-tuned variant of Qwen2.5-Math-7B and supports token-level log-probability outputs.

For model interchange experiments, we select two baseline models and two continuation models:

\subsection{Baseline}

To establish a baseline for reasoning quality, we employ Gemma-3-4B-IT and LLaMA-3.1-70B-Instruct to generate CoT exemplars, two state-of-the-art instruction-tuned models from distinct architectural lineages.

\begin{itemize}
    \item Gemma-3-4B-IT \citep{gemmateam2025gemma3technicalreport}, an instruction-tuned variant from the Gemma 3 model family developed by Google Deepmind with 4 Billion parameters is used to generate complete Chain-of-Thought reasoning paths, tuned to Gemma's architecture.
\end{itemize}
\begin{itemize}
    \item LLaMA-3.1-70B-Instruct \citep{grattafiori2024llama3herdmodels}, a large scale variant from the LLaMA 3 model family developed by Meta AI with 70 Billion parameters is used to generate complete Chain-of-Thought reasoning paths, tuned to LLaMA's architecture.
\end{itemize}

\subsection{Continuation}
\begin{itemize}
    \item Gemma-3-1B-IT \citep{gemmateam2025gemma3technicalreport}, a lightweight variant from the same Gemma 3 model family, is used to evaluate how well reasoning chains can be completed by a structurally similar but smaller model.
\end{itemize}
\begin{itemize}
    \item LLaMA 3.1-8B-Instruct \cite{grattafiori2024llama3herdmodels} representing a different architectural lineage helps enable testing interchangeability across distinct LLM families. For brevity, we refer to the aforementioned models as Gemma and LLaMA respectively for the remainder of this paper.

\end{itemize}

On the MATH dataset, Gemma 3-1B-IT and Gemma 3-4B-IT performed with accuracies of $48.0\%$ and $75.6\%$ respectively \citep{gemmateam2025gemma3technicalreport}. 
Moreover, Llama-3.1-8B-Instruct and Llama-3.1-70B-Instruct performed with accuracies $47.2\%$ and $65.7\%$ \citep{yang2024qwen25mathtechnicalreportmathematical}. 
We observe that the two base models exhibit similar performance levels and, likewise, that the two continuation models perform comparably.

All models were prompted using one consistent CoT templates, either the interchange or full-run variant, as detailed in \hyperref[app:prompting]{Section~\ref*{app:prompting}}.

\subsection{Evaluation Metrics}
\label{sec:eval-metrics}
The  hybrid reasoning chains generated were evaluated using a multifaceted set of metrics designed to assess accuracy, variability, and the impact of model interchanges on reasoning coherence and final outcomes. Specifically, we consider the following four core metrics:

\begin{itemize}
    \item Answer Accuracy: Accuracy is defined as the proportion of final answers from generation that exactly match those from ground-truth solutions. This metric represents the model's ability to arrive at the correct final result through its reasoning chain.

    \item PRM Score: As a PRM is available for scoring, we additionally report average PRM-assigned scores that capture the internal likelihood and coherence of a given chain regardless of final correctness. 
We define the PRM score \( A' \) as the average plausibility score assigned to each reasoning step in a chain of \( n \) steps:

    \[
    A' = \frac{1}{n} \sum_{i=1}^{n} \text{PRM}(s_i)
    \]

    where \( s_i \) denotes the \( i \)-th step in the chain. While traditional accuracy reflects outcome-level correctness, \( A' \) provides a step-level assessment of reasoning quality.
\end{itemize}

\begin{itemize}
    \item Normalized Relative Gain (NRG): This metric quantifies whether incorporation of reasoning from another model helps or hinders performance. Given the accuracies of the original model \( A \) and \( B \), and hybrid accuracies \( A' \) (Model A prefix + Model B suffix) and \( B' \) (Model B prefix + Model A suffix), we define:

{\small
\[
\text{NRG}_A = \frac{A' - A}{A}, \quad
\text{NRG}_B = \frac{B' - B}{B}.
\]
}

Positive values indicate a performance gain from model interchange, while negative values reflect degradation.
\end{itemize}
\begin{itemize}
    \item Cross-Model Degradation (XMD): This metric captures the extent to which the continuation of a model degrades the original reasoning trajectory. It is defined as:

{\small
\[
\text{XMD}_{A \to B} = \frac{A - B'}{A}, \quad
\text{XMD}_{B \to A} = \frac{B - A'}{B}.
\]
}

XMD provides a normalized measure of reasoning incompatibility, where higher values indicate more severe disruptions introduced by the cross-model continuation.
\end{itemize}

\section{Results}
\label{sec:results}
We present results across the MATH benchmark to evaluate model interchangeability across truncation points. Through our proposed metrics, we look to determine whether model continuation works to improve or disrupt the original reasoning trajectory. Experimental results were obtained using the Runpod cloud platform, leveraging NVIDIA H100 PCIe GPUs over approximately 250 GPU hours.

\subsection{Full Chain-of-Thought Results}

To establish a baseline, we first evaluate each model’s performance using end-to-end CoT reasoning applied without interruption. For every example in the benchmark, the model is prompted to reason step by step to completion, producing a complete trajectory from question to final answer. We report results in terms of final answer accuracy and step-level reasoning score as seen in Table~\ref{tab:interchangeability-combined}.

This baseline allows us to quantify native reasoning strengths and weaknesses of each model without the effects of interchange.

\begin{table}[H]
  \vspace{1em}
  \centering
  {\small  
  \begin{tabular}{lccc}
    \toprule
    \textbf{Model} & \textbf{Dataset} & \textbf{Accuracy (\%)} & \textbf{PRM} \\
    \midrule
    Gemma-3-4B-IT & MATH & 68.06\%& 0.8952 \\
    Gemma-3-1B-IT & MATH & 36.28\%& 0.7904\\
    LLaMA-3.1-70B-Instruct & MATH & 60.80\%& 0.8725 \\
    LLaMA-3.1-8b-Instruct & MATH & 47.76\%&  0.8522 \\
    \bottomrule
  \end{tabular}
  }
  \vspace{1em}
  \caption{Performance of reasoning chains fully generated by each model (i.e., with no handoff or interchange from another model) on the MATH dataset.}
  \label{tab:interchangeability-combined}
\end{table}

\subsection{Interchanged Chain-of-Thought Results}

Thereafter, to gauge the interchangeability of reasoning processes across different models, we evaluate the completion of truncated CoT traces. Each reasoning chain is strategically truncated based on cumulative log-probability thresholds $(25\%, 50\%, 75\%)$, representing early, mid, and late points in the reasoning process. Subsequently, alternative models are assigned to continue the truncated reasoning chains through to completion.

We report performance for all continuation combinations, including accuracy, step-level scores, and coherence ratings as seen in Table~\ref{tab:interchangeability-math-gemma} \& Table~\ref{tab:interchangeability-math-llama}. This analysis unveils the extent to which partial reasoning from one model can be reliably extended by another, highlighting cases of both successful handoff and systematic breakdowns that point to the limits of reasoning interchangeability.

\begin{table}[H]
\centering
\begin{tabular}{llllcc}
\toprule
\textbf{Truncation} & \textbf{Continuation} & \textbf{Accuracy (\%)} & \textbf{PRM} & \textbf{NRG} & \textbf{XMD} \\
\midrule
25\% & Gemma-3-1B-IT & 41.76\% & 0.7966 & 0.3678 & 0.3864 \\
25\% & LLaMA-3.1-8B-Instruct & 43.60\% & 0.8393 & 0.3196 & 0.3594 \\
50\% & Gemma-3-1B-IT & 49.86\% & 0.8002 & 0.3786 & 0.2674 \\
50\% & LLaMA-3.1-8B-Instruct & 53.24\% & 0.8585 & 0.264 & 0.2177 \\
75\% & Gemma-3-1B-IT & 55.26\% & 0.8032 & 0.3500 & 0.1881 \\
75\% & LLaMA-3.1-8B--Instruct & 63.80\% & 0.8697 & 0.1853 & 0.0626 \\
\bottomrule
\end{tabular}%
\vspace{1em}
\caption{Performance of hybrid reasoning chains by truncation point and continuation model on MATH dataset, using a fully generated CoT from Gemma-3-4B-IT.}
\label{tab:interchangeability-math-gemma}
\end{table}

\begin{table}[H]
\vspace{1em}
\centering
\begin{tabular}{llllcc}
\toprule
\textbf{Truncation} & \textbf{Continuation} & \textbf{Accuracy (\%)} & \textbf{PRM} & \textbf{NRG} & \textbf{XMD} \\
\midrule
25\% & Gemma-3-1B-IT& 36.16\% & 0.7566 & -0.1137 & 0.4053 \\
25\% & LLaMA-3.1-8B-Instruct& 42.18\% & 0.8323 & -0.0150 & 0.3062 \\
50\% & Gemma-3-1B-IT& 38.50\% & 0.7730 & -0.0968 & 0.3668 \\
50\% & LLaMA-3.1-8B-Instruct& 46.26\% & 0.8456 & -0.0072 & 0.2391 \\
75\% & Gemma-3-1B-IT& 41.98\% & 0.7811 & -0.0827 & 0.3095 \\
75\% & LLaMA-3.1-8B-Instruct& 50.06\% & 0.8543 & -0.0002 & 0.1766 \\
\bottomrule
\end{tabular}%
\vspace{1em}
\caption{Performance of hybrid reasoning chains by truncation point and continuation model on MATH dataset, using a fully generated CoT from LLaMA-3.1-70B-Instruct.}
\label{tab:interchangeability-math-llama}
\end{table}



\section{Discussion}
\label{sec:discussion}
Our observations uncover degradation in performance when cross-family models are tasked to continue reasoning midstream initiated by a partially completed CoT. There are several factors likely responsible for this downgrade in performance: 

\subsection{Style and Representational Compatibility}
A consistent disparity between intra-family and cross-family continuation highlights representational compatibility as a key factor in multi-model reasoning. Despite receiving high confidence chains, cross-family continuations (e.g., Gemma-3-4B-IT→LLaMA-3.1-8B-Instruct and LLaMA-3.1-70B-Instruct→Gemma-3-1B-IT) often fail to maintain correct reasoning. For instance, when LLaMA-3.1-70B-Instruct’s chain is continued by Gemma-3-1B-IT, accuracy falls to $36.16\%$ at the $25\%$ mark-nearly a 40\% relative decline compared to the base model’s $60.80\%$ full-chain accuracy-with a corresponding negative NRG ($-0.1137$). Similarly, continuations from Gemma-3-4B-IT into LLaMA-3.1-8B-Instruct under perform early on ($43.60\%$ at $25\%$) despite access to confident reasoning prefixes, yielding a lower NRG of $0.3196$ compared to intra-family continuation at the same depth (Gemma-3-4B-IT→Gemma-3-1B-IT, $0.3678$), indicating that these prefixes do not fully overcome differences in architecture and reasoning style. This pattern suggests a reasoning bias: each model family tends to rely more on its own reasoning patterns, which may result from structural differences between the families.


These results are consistent with previous work \citep{liu2023logicot}, which noted that structural differences between model families (GPT-4 in their case) can limit cross-model reasoning transfer, particularly for complex, multi-step reasoning tasks. While LLaMA models generate coherent chains within their own family, their internal reasoning representations differ from Gemma’s, which may hinder smooth continuation across families. This is supported by consistently high XMD values across truncation points (e.g., $0.4053$ at $25\%$ and $0.3095$ at $75\%$ for LLaMA-3.1-70B-Instruct→Gemma-3-1B-IT), suggesting that reasoning coherence is not fully maintained even as longer prefixes are available. High-confidence reasoning prefixes do not appear sufficient to completely navigate these differences, indicating that cross-family continuation is constrained by family-specific reasoning tendencies. 


In contrast, intra-family continuations show steady improvement with longer truncation depths. For example, when Gemma-3-1B-IT continues from Gemma-3-4B-IT, accuracy rises from $41.76\%$ at $25\%$ to $55.26\%$ at $75\%$, accompanied by moderate NRG values ($0.3678$→$0.3500$) and decreasing XMD ($0.3864$→$0.1881$). Similarly when LLaMA-3.1-8B-Instruct continues from LLaMA-3.1-70B-Instruct, performance increases from $42.18\%$ to $50.06\%$, with NRG improving from $-0.0150$ to near-neutral ($-0.0002$) and XMD decreasing from $0.3062$ to $0.1766$. These patterns suggest that the representational similarity between models supports a more stable continuation and better integration of context.

\subsection{Context Integration Overhead}
When deployed late in the reasoning chain (e.g., at the $75\%$ mark), smaller continuation models such as Gemma-3-1B-IT and LLaMA-3.1-8B-Instruct must interpret and integrate extensive context generated by larger base models (Gemma-3-4B-IT and LLaMA-3.1-70B-Instruct). As reasoning sequences lengthen, models may face capacity limits that degrade performance. This bottleneck is attributed to the finite “working memory” of LLMs and the compounding demands of maintaining logical coherence across many steps \citep{unitedminds}. The effect is especially pronounced when models are required to interpret and continue reasoning from an externally provided chain rather than generating all steps from scratch.

    On the MATH dataset, truncation depth produces gradual improvements but does not eliminate the performance gap relative to non-handoff baselines. For example, when continuing Gemma-3-4B-IT’s reasoning, Gemma-3-1B-IT improves from $41.76\%$ at $25\%$ truncation to $55.26\%$ at $75\%$, while LLaMA-3.1-8B-Instruct rises from $43.60\%$ to $63.80\%$. Similarly, when continuing LLaMA-3.1-70B-Instruct, LLaMA-3.1-8B-Instruct achieves a smoother progression from $42.18\%$ to $50.06\%$, outperforming Gemma-3-1B-IT, which remains between $36.16\%$ and $41.98\%$. These trends suggest that architectural alignment facilitates smoother context integration in same-family continuations, while representational mismatches in cross-family pairs disrupt coherent reasoning.

    Despite improvements with longer prefixes, performance remains notably below that of fully self-generated chains (Gemma-3-4B-IT: $68.06\%$, LLaMA-3.1-70B-Instruct: $60.80\%$). XMD values confirm this persistent overhead: even at the $75\%$ truncation point, XMD remains non-negligible ($0.0626$ for Gemma-3-4b-IT→LLaMA-3.1-8B-Instruct and $0.1766$ for LLaMA-3.1-70B-Instruct→LLaMA-3.1-8B-Instruct), indicating incomplete recovery of original reasoning quality.

    These observations highlight that truncation depth alone does not ensure effective reasoning transfer. Although larger prefixes reduce uncertainty and contextual loss, architectural and stylistic compatibility between base and continuation models remains the key factor determining success.

\subsection{Error Amplification}
Minor inconsistencies or ambiguities in early reasoning steps, especially when generated by a different model, can compound as LLaMA or Gemma continue the reasoning process. With limited steps remaining to revise earlier logic (particularly in final-answer-only completions), both models struggle to recover from upstream errors. These results suggest that effective interoperability in multi-step reasoning depends on both model capability and the degree of representational and contextual alignment across reasoning steps.


    On the MATH dataset, when reasoning chains generated by Gemma-3-4B-IT or LLaMA-3.1-70B-Instruct are truncated and continued by smaller models at various points ($25\%, 50\%, 75\%$), performance declines in proportion to both truncation depth and cross-family divergence. When Gemma-3-4B-IT serves as the base, continuation by Gemma-3-1B-IT (intra-family) improves steadily from $41.76\%$ at $25\%$ to $55.26\%$ at $75\%$, with NRG values rising from $0.3678$ to $0.3500$ and XMD decreasing from $0.3864$ to $0.1881$. Cross-family continuation by LLaMA-3.1-8B-Instruct performs competitively ($43.60\%$→$63.80\%$) but shows slightly lower NRG ($0.3196$→$0.1853$), indicating weaker efficiency in utilizing the provided context. Longer prefixes appear to partially reduce representational mismatch, leading to more consistent performance over time.
    

    When LLaMA-3.1-70B-Instruct serves as the base model, Gemma-3-1B-IT continuations perform substantially worse ($36.16–41.98\%$ across truncation points) with persistently high XMD ($0.4053$→$0.3095$) and negative NRG ($-0.1137$→$-0.0827$), suggesting limited transfer across families. Intra-family continuation by LLaMA-3.1-8B-Instruct performs more stably, reaching $50.06\%$ at $75\%$ with NRG improving from $-0.0150$ to $-0.0002$ and XMD decreasing from $0.3062$ to $0.1766$, reflecting more consistent reasoning integration within the same family.
    

    Comparing fully generated chains with the hybrid results (Table \ref{tab:interchangeability-combined}), Gemma-3-4b and LLaMA-3.1-70B-Instruct still substantially outperform their continuations ($68.06\%$ and $60.80\%$, respectively). However, their smaller counterparts, especially Gemma-3-1B-IT, demonstrate partial to considerable recovery when inheriting sufficiently long prefixes, suggesting that similar architecture and tokenization structures enhance transfer performance. As seen over $25\%/50\%/75\%$ truncations, intra-family continuation (Gemma-3-4b→Gemma-3-1b) improves from $41.76\%$→$55.26\%$ ($+13.5$ pp), and even cross-family continuation (Gemma-3-4B-IT→LLaMA-3.1-8B-Instruct) exhibits a greater net gain of $43.60\%$→$63.80\%$ ($+20.2$ pp). In contrast, continuations from LLaMA-3.1-70b displayed weaker recovery with LLaMA-3.1-8B-Instruct rising only $+7.9$ pp ($42.18\%$
    $50.06\%$), and Gemma-3-1B-IT gains just $+5.8$ pp ($36.16\%$→$41.98\%$). As truncation length increases, reasoning becomes more coherent, but full recovery is still unattainable, lending credence to how small representational gaps can compound through multi-step reasoning chains.

\section{Conclusion}

In this work, we introduced a novel framework for evaluating midstream interchangeability in large language models, grounded in a chain-splitting paradigm based on cumulative log-probability. By systematically truncating the reasoning chains generated by our base models and appending completions from either intra-family or cross-family models, we directly measured the stability and coherence of hybrid reasoning trajectories. Our experiments on MATH demonstrate that model family alignment plays a decisive role in the success or failure of such hybrid chains. While intra-family continuations generally preserved reasoning quality on simpler tasks, cross-family continuations often struggled to maintain coherence with the partial chains, despite comparable model performance as referenced in \hyperref[sec:model-selection]{Section~\ref*{sec:model-selection}}. This suggests that models like Gemma and LLaMA may be better aligned to continue reasoning within their own family than across different architectures.

These findings challenge previous assumptions about model modularity in contemporary NLP. Despite architectural advances and increasing performance parity across model families, our results suggest that inter-model transfer in multi-step reasoning remains fragile, constrained by differences in stylistic alignment, latent variable encoding, and contextual integration. The observed breakdowns reveal a significant gap between individual task performance and interoperability in reasoning, which is an area that has received insufficient attention in LLM evaluation.

More broadly, our work highlights the need for new approaches that preserve consistent semantic reasoning across different language models. As research advances toward compositional and multi-agent LLM systems, reliable interchangeability will become essential, not solely for efficiency, but also for alignment, verification, and interpretability. Our methodology provides an initial framework for diagnosing and quantifying this interoperability gap in a systematic, data-driven manner.


\section*{Limitations}
\label{sec:limitations}

\begin{itemize}
    \item Single Completion Runs: All experiments were conducted using deterministic continuations. While this reflects realistic deployment scenarios, it limits our understanding of variance under sampling. Future work should evaluate robustness using multiple stochastic rollouts.
\end{itemize}
\begin{itemize}
    \item Task Domain Scope: Our evaluation is confined to math reasoning (MATH). It remains unclear whether interchangeability generalizes to commonsense, scientific, or multimodal reasoning tasks.
\end{itemize}
\begin{itemize}
    \item Domain-Specific PRMs: We employed a math-specific Process Reward Model (PRM). Evaluating reasoning quality in other domains will require retraining or adapting PRMs tailored to those reasoning distributions.
\end{itemize}

\section*{Future Work}

\begin{itemize}
\raggedright
    \item Cross-Domain Generalization: Evaluate model interchangeability on tasks such as commonsense QA, multi-hop retrieval, scientific explanation, and instruction-following, where reasoning formats may be more variable or implicit.
    \item Adaptive Truncation Strategies: Rather than using static log-probability thresholds $(25/50/75\%)$, future work could explore dynamic segmentation based on reasoning content, semantic shifts, or model uncertainty.
    \item Collaborative Model Architectures: Deploy multi-agent or multi-model reasoning pipelines in production environments (e.g., tutoring systems, scientific assistants) to study tradeoffs in latency, memory, and correctness.
\end{itemize}

\section{Appendix}
\subsection{Prompting}
\label{app:prompting}
\begin{promptbox}{Standardized Prompt}
\textbf{Full-Run Prompt 
}: \\\\
\textit{System message:}
"You are a helpful assistant that solves problems step by step.
Please provide clear reasoning with numbered steps and conclude with your final answer."

\vspace{6pt}
\textit{User message:} "Solve this problem step by step:

Question: ['question']"\\

\textbf{Interchange Prompt 
}: \\\\
\textit{System message:}
"You are a helpful assistant that solves problems step by step.
Please provide clear reasoning with numbered steps and conclude with your final answer."

\vspace{6pt}
\textit{User message:} "Solve this problem step by step:

Question: ['question']
['truncated reasoning']"

\end{promptbox}
This prompt standardization ensures comparability in reasoning styles across models; slight variations were applied where necessary to accommodate model-specific tokenization or formatting requirements without altering the intended instructions or task semantics.

\bibliographystyle{plainnat} 
\bibliography{custom}

\end{document}